\documentclass[letterpaper, 10 pt, conference]{ieeeconf}  

\IEEEoverridecommandlockouts                              

\usepackage{graphics} 
\usepackage{epsfig} 
\usepackage{mathptmx} 
\usepackage{times} 
\usepackage{amsmath} 
\usepackage{amssymb}  
\usepackage{cite}
\usepackage{amsfonts}

\usepackage{helvet}
\usepackage{courier}
\usepackage{color}

\usepackage{algorithmic}
\usepackage{algorithm}
\usepackage{graphicx}
\usepackage{textcomp}
\usepackage{xcolor}
\usepackage{multirow}
\usepackage{float}
\usepackage{bm}
\usepackage{caption}
\usepackage{soul}
\usepackage{url}
\usepackage{cool}
\usepackage{nameref}
\usepackage{float}
\usepackage[utf8x]{inputenc} 
\usepackage[breaklinks=true,bookmarks=false]{hyperref}
\usepackage{textcomp}

\title{\LARGE \bf Adaptive-SpikeNet: Event-based Optical Flow Estimation using Spiking Neural Networks with Learnable Neuronal Dynamics}

\author{Adarsh Kumar Kosta and Kaushik Roy 
\thanks{All authors are with Purdue University, West Lafayette, IN 47907, USA {\tt\small \{akosta, kaushik\}@purdue.edu}}
}

\begin{document}

\newcommand{\ignore}[1]{}

\newif\ifsubmit
\submitfalse
\ifsubmit
    \newcommand{\akosta}[1]{}
    \newcommand{\kaushik}[1]{}
    \newcommand{\todo}[1]{}
    \newcommand{\tocite}[1]{}
    \newcommand{\update}[1]{}
\else
    \newcommand{\akosta}[1]{[{\color{brown}AK: #1}]}
    \newcommand{\kaushik}[1]{[{\color{cyan}KR: #1}]}
    \newcommand{\todo}[1]{[{\color{red}TODO: #1}]}
    \newcommand{\tocite}[1]{[{\color{red}CITE: #1}]}
    \newcommand{\update}[1]{{\color{brown} #1}}
\fi

\maketitle
\thispagestyle{empty}
\pagestyle{empty}

\begin{abstract}
Event-based cameras have recently shown great potential for high-speed motion estimation owing to their ability to capture temporally rich information asynchronously. Spiking Neural Networks (SNNs), with their neuro-inspired event-driven processing can efficiently handle such asynchronous data, while neuron models such as the leaky-integrate and fire (LIF) can keep track of the quintessential timing information contained in the inputs. SNNs achieve this by maintaining a dynamic state in the neuron memory, retaining important information while forgetting redundant data over time. Thus, we posit that SNNs would allow for better performance on sequential regression tasks compared to similarly sized Analog Neural Networks (ANNs). However, deep SNNs are difficult to train due to vanishing spikes at later layers. To that effect, we propose an adaptive fully-spiking framework with learnable neuronal dynamics to alleviate the spike vanishing problem. We utilize surrogate gradient-based backpropagation through time (BPTT) to train our deep SNNs from scratch. We validate our approach for the task of optical flow estimation on the Multi-Vehicle Stereo Event-Camera (MVSEC) dataset and the DSEC-Flow dataset. Our experiments on these datasets show an average reduction of $\sim13\%$ in average endpoint error (AEE) compared to state-of-the-art ANNs. We also explore several down-scaled models and observe that our SNN models consistently outperform similarly sized ANNs offering $\sim10\%$-$16\%$ lower AEE. These results demonstrate the importance of SNNs for smaller models and their suitability at the edge. In terms of efficiency, our SNNs offer substantial savings in network parameters ($\sim48.3\times$) and computational energy ($\sim10.2\times$) while attaining  $\sim10\%$ lower EPE compared to the state-of-the-art ANN implementations.
\end{abstract}

\section{INTRODUCTION} \label{sec:intro}
Research in the fields of neuroscience, machine learning and robotics has been highly inspired by the behaviour of biological species in their natural environments. Be it a house-fly seamlessly navigating through cluttered spaces, an osprey diving to catch a fish, a squirrel jumping from one branch to the other or a bee precisely landing on a swaying flower~\cite{flymotion2010, honeybee1996, baird2013universal, serres2017optic}. All these behaviours require a comprehensive understanding of the environment dynamics at an extremely low-latency and energy. Visualizing the motion field in such environments, often referred to as optical flow is one of the fundamental tasks involved. The ability to perform accurate and low-latency optical estimations would greatly benefit modern day artificial intelligence (AI)-powered robots such as drones, ground robots, and autonomous vehicles. It will further fuel more complex tasks such as obstacle detection/avoidance and path planning, enabling seamless autonomous navigation.  Modern AI aims to achieve this using data from sensors such as frame-based cameras coupled with Analog Neural Networks (ANNs\footnote[1]{We refer to deep learning networks as ANNs due to their analog nature of inputs/computations, even though the underlying computing hardware can be digital.}). However, such models when optimized for performance turn out to be compute and memory intensive while having a high-latency. On the other hand, smaller models suitable for the edge suffer in terms of performance and require constant communication with the cloud.
In light of this, there is a need to deliberate over our outlook towards tackling such tasks and rethink the processing pipeline all the way from sensors, to algorithms, to underlying hardware.

For sequential regression tasks with rapidly changing inputs, the timing information between the inputs is crucial and needs to be captured at a high temporal resolution. Frame-based cameras fail in such scenarios due to their fixed and low temporal resolution. They also suffer from motion blur, temporal aliasing and poor image quality in low light and high dynamic range scenes. Event-based cameras, such as the Dynamic Vision Sensor (DVS)~\cite{dvs128, dvs240}, overcome these issues by asynchronously sampling the log intensity changes at every pixel independently. They promise a high temporal resolution in the order of microseconds, a high dynamic range (140dB vs 60dB) and extremely low power consumption (10mW vs 3W) compared to frame-based cameras. However, ANN-based methods which were originally designed for frame-based images turn out to be incompatible at directly handling the binary and asynchronous outputs generated by event cameras. ANNs discard the temporal ordering of inputs by representing them as channels and perform sub-optimal stateless computations.

Spiking Neural Networks (SNNs), often referred to as the third generation of neural networks, offer asynchronous event-driven compute that fits naturally with event data. SNNs perform computations only at the arrival of inputs generating progressively sparser outputs as the network depth increases. Due to the implicit recurrence, SNNs can preserve the input timing information in the neuronal state called `membrane potential' based on parameters such as neruon firing threshold and leak. This makes them inherently suitable for handling sequential tasks efficiently. SNNs implemented on neuromorphic processors such as Loihi~\cite{loihi2018} and TrueNorth~\cite{truenorth2015} can offer several orders of magnitude lower energy consumption and latency compared to traditional von Neumann architectures.
Over the past few years, direct SNN training for event-based vision has been limited to small scale and discrete problems such as image classification and action recognition on simplistic datasets like N-MNIST~\cite{orchard2015converting}, DVS128 Gesture~\cite{amir2017low} etc. This is due to the fact that deep SNNs are difficult to train, mainly owing to vanishing spikes at deeper layers.

To that effect, we propose an adaptive fully-spiking framework with learnable neuronal dynamics for the complex regression task of optical flow estimation. Our proposed framework overcomes the spike vanishing problem in SNNs, effectively capturing the temporal information contained in the inputs and allowing training of deep SNNs. Our main contributions are as follows:


\begin{itemize}
\item We propose a fully-spiking framework with adaptive leaky-integrate and fire (LIF) neurons that can be trained from scratch using surrogate gradients during backpropagation, for the sequential regression task of event-based optical flow estimation.
\item We show that our fully-spiking models outperform state-of-the-art ANN~\cite{zhu2018ev} and hybrid SNN-ANN~\cite{lee2020_spikeflownet} models, both in terms of performance and compute efficiency on MVSEC~\cite{zhu2018multivehicle} and DSEC-Flow~\cite{gehrig2021dsec, gehrig2021raft} datasets.
\item We demonstrate the importance of effectively capturing temporal information by analyzing various model sizes and observe that the performance difference between correspondingly sized SNNs and ANNs remains consistent upon reducing the model size. 
\end{itemize}

\section{Related Works} \label{sec:related_works}
\vspace{-1mm}
With event cameras showcasing great potential for low-latency optical flow estimation, there have been several (optimization and non-learning based) works in the recent years exploring this research area~\cite{benosman2, brosch2015event, gallego2018}.
In learning-based ANN methods, the first ANN employed self-supervised learning (SSL) for event-based flow estimation and was proposed in~\cite{zhu2018ev} inspired by the U-Net~\cite{unet} architecture. Another similar approach was presented in ~\cite{ye2020_unsupervised} which predicted optical flow using depth and pose estimates. These approaches used separate channels to provide encoded timing information as inputs to the ANN model which was sub-optimal. Researchers in ~\cite{zhu2019unsupervised}, introduced a voxel-based event-input representation inspired by ~\cite{benosman2} and a contrast maximization loss~\cite{gallego2018} removing the dependency on grayscale images for loss computations. Authors in~\cite{paredes2021back} extend this to image reconstruction with events and propose a new lightweight architecture called Fire-FlowNet for flow estimation. In contrast to the above SSL pipelines, authors in ~\cite{stoffregen2020reducing} utilized synthetic data from an event simulator ~\cite{rebecq2018esim} to train the networks proposed in ~\cite{zhu2018ev, zhu2019unsupervised} and obtained significantly better performance. However, all these works do not specifically focus on exploiting the timing information or temporal ordering between inputs. A recent supervised work called E-RAFT~\cite{gehrig2021raft} explored the importance of timing information by explicitly using temporal history and leveraging correlation volumes for iterative flow refinement.

Regarding learning-based SNN works for optical flow estimation, an STDP (Spike Timing Dependent Plasticity) based convolutional SNN was presented in ~\cite{paredes2019unsupervised} but was limited to relatively simplistic flow behaviour. Following this, authors in ~\cite{lee2020_spikeflownet} presented a hybrid SNN-ANN version of ~\cite{zhu2018ev} with a SNN encoder and ANN decoder, outperforming it and establishing the utility of SNNs. Researchers in~\cite{lee2022fusion} expanded this framework by performing sensor fusion and adding a secondary ANN-based encoder for frame-based data.
In purely SNN-based approaches, researchers in ~\cite{hagenaars2021self} explored parameter initialization, surrogate gradients and adaptive neuronal mechanisms for flow estimation. However, their results were inconclusive in several scenarios for the explored neuronal mechanisms. Although, we agree that adding explicit recurrence improves performance as pointed out in ~\cite{gehrig2021raft, hagenaars2021self},  we believe that relaxing the constraints on the adaptivity of neuronal dynamics and exploiting the implicit recurrence offered by SNNs can lead to better performance without additional parameters and learning overheads. Works such as ~\cite{rathi2021diet, fang2021incorporating, yin2020effective} show the importance of neuronal dynamics towards improving performance, however, they target more straightforward classification tasks. We explore their extendability to a complex regression task of optical flow estimation through this work.

\section{Method} \label{sec:method}
\subsection{Sensors and Input representation} \label{subsec:sensors}

\vspace{-4mm}
\begin{figure}[h]
\begin{center}
\includegraphics[width=0.48\textwidth]{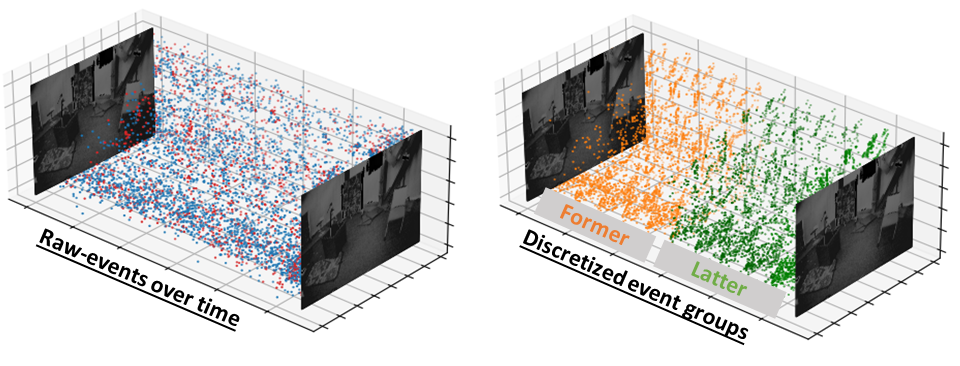}
\vspace{-4mm}
\caption{($left$) Raw event stream between two consecutive frames. ($right$) Event bins in former and latter event groups.}
\vspace{-6mm}
\label{fig:inp_rep}
\end{center}
\end{figure}
Event-based cameras, inspired by the biological retina ~\cite{mahowald1994silicon}, sample the log intensity changes at each discrete pixel asynchronously and independently. Any change in the log intensity ($I$) over a specified threshold ($\theta$) is recorded as a discrete event at that pixel location, (i.e., $\|\log(I_{t+1}) - \log(I_{t})\| \geq \theta$). The data is generated in the Address Event Representation (AER) format comprising a tuple $\{x, y, t, p\}$, with $(x,y)$ representing pixel locations, ($t$) representing the camera timestamp and ($p$) representing the (ON/OFF) polarity of the intensity change.

In this work, we utilize the discretized event volume representation as presented in ~\cite{zhu2019unsupervised}. For a set of $N$ input events $\{(x_i, y_i, t_i, p_i)\}_{i\in[1, N]}$ between two consecutive grayscale images and a set of $B$ event bins to be created within this event volume, the discretized event volume is generated using bilinear sampling as follows:
\begin{equation}
    t_i^* = (B-1)(t_i-t_1)/(t_N-t_1) \\ \tag{1}
\end{equation}
\begin{equation}
    V(x,y,t) = \sum_i{p_i k_b (x-x_i) k_b(y-y_i) k_b(t-t_i^*)} \\ \tag{2}
\end{equation}
\begin{equation}
    k_b = max[0, 1-|a|]\\ \tag{3}
\end{equation}
where $k_b(a)$ is the bilinear sampling kernel from ~\cite{jaderberg2015}. We further process the obtained discretized voxel into former and latter groups of event bins corresponding to each (ON/OFF) polarity leading to a 4-channeled representation involving former and latter event groups, as described in ~\cite{lee2020_spikeflownet}. These groups facilitate flow estimation between equidistant bins ($5$) to obtain as many flow predictions as possible in the event volume while avoiding different temporal scales. Each channel thus contains a total of $T=B/2$ event frames that are passed sequentially as timesteps through the network. The idea behind passing a multi-channel representation at each timestep is to allow the network to learn a larger temporal correlation directly while capturing the short term temporal correlation in neuronal dynamics over timesteps. This representation preserves the spatio-temporal information in the event stream while offering high energy-efficiency due to controlled number of timesteps ($T$) for sequential processing. Fig.~\ref{fig:inp_rep} shows the former and latter event groups based event input representation.

\begin{figure}[ht]
\begin{center}
\includegraphics[width=0.48\textwidth]{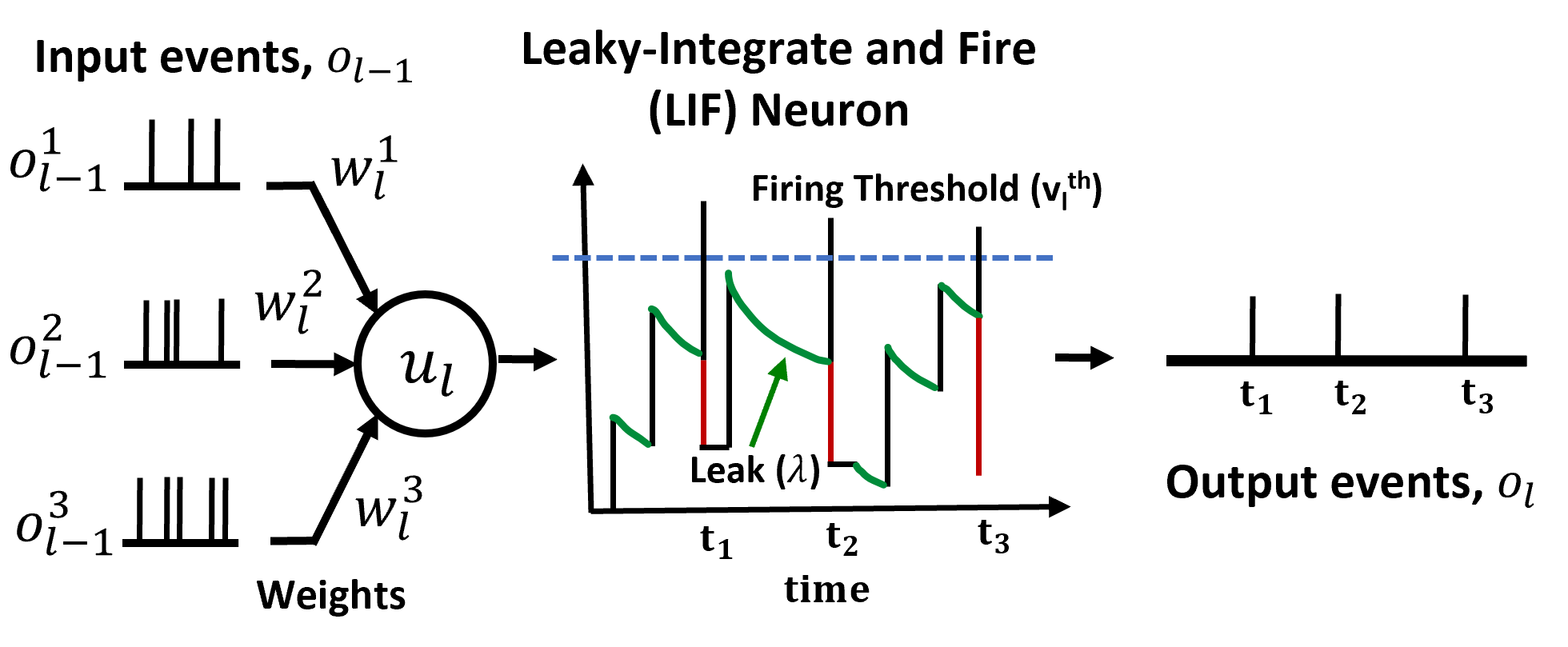}
\vspace{-4mm}
\caption{Leaky-Integrate and Fire (LIF) neuron. The firing threshold ($v^{th}_l$) and leak factor ($\lambda$) are learnable parameters.}
\label{fig:lif_dynamics}
\vspace{-5mm}
\end{center}
\end{figure}

\subsection{Neuron Model} \label{subsec:neuron_models}
There exist several biologically inspired neuron models of which the Leaky-Integrate-and-Fire (LIF)~\cite{abbott1999lapicque} is the most widely used due to its immense capability of storing and recalling information, yet being simple enough to not explode the model parameters. The LIF neuron allows to accumulate information over time into the neuronal state called `membrane potential' ($u$), while also enabling controlled forgetting over time (through leak, $\lambda$). The LIF neuron model in an SNN can be described as:
\begin{equation}
    \boldsymbol{u}_l^t = \lambda_l\boldsymbol{u}_l^{t-1}  +  \boldsymbol{W}_lo_{l-1}^t - v^{th}_l \boldsymbol{o}_l^{t-1} \\ \tag{4}
    \label{eq4}
\end{equation}
where $ \boldsymbol{u}_l^t$ represents the membrane potential of the  layer $l$ at timestep $t$, $ \boldsymbol{W}_l$ represents the synaptic weights connecting layers $l-1$ and $l$, $o_l$ represents the binary output spike at layer $l$, $v^{th}_l$ is the neuronal firing threshold and $\lambda_l$ represents the neuronal leak factor for layer $l$. The first term in Eq.\ref{eq4} denotes the leakage in membrane potential, the second term computes the summation of weighted output spikes from the previous layer, and the third term denotes the reduction of membrane potential upon generation of an output spike at the current layer. This reduction by an amount equal to the firing threshold ($v^{th}_i$) is termed as `soft reset', while a reset to zero is termed as `hard reset'.
The generation of output spikes follows the following equation at each timestep:
\begin{equation}
     \boldsymbol{z}_l^{t} =  \boldsymbol{u}_l^{t}/v^{th}_l - 1, \; \; \; \;  
     \boldsymbol{o}_l^{t} =\begin{cases}
               1, & \text{if~$ \boldsymbol{z}_l^{t}>0$}\\
               0, & \text{otherwise}
            \end{cases} \tag{5}
    \label{eq5}
\end{equation}

These operations are carried out over all timesteps in each neuron enabling event-driven computations. The threshold governs the average integration time of inputs while the leak controls the amount of membrane potential retained across timesteps. Traditionally, the neuronal firing threshold ($v^{th}$) and the leak factor ($\lambda$) are fixed {\em a priori}. We posit that making $v^{th}$ and $\lambda$ parameters trainable would allow the network to maintain sufficient spiking activity eliminating vanishing spikes and remember temporal information using the membrane potential, allowing deep SNNs to learn complex tasks ~\cite{rathi2021diet}. Fig~\ref{fig:lif_dynamics} shows the LIF neuron dynamics.

\begin{figure*}[ht]
\vspace{1.5mm}
\begin{center}
\includegraphics[width=0.84\textwidth]{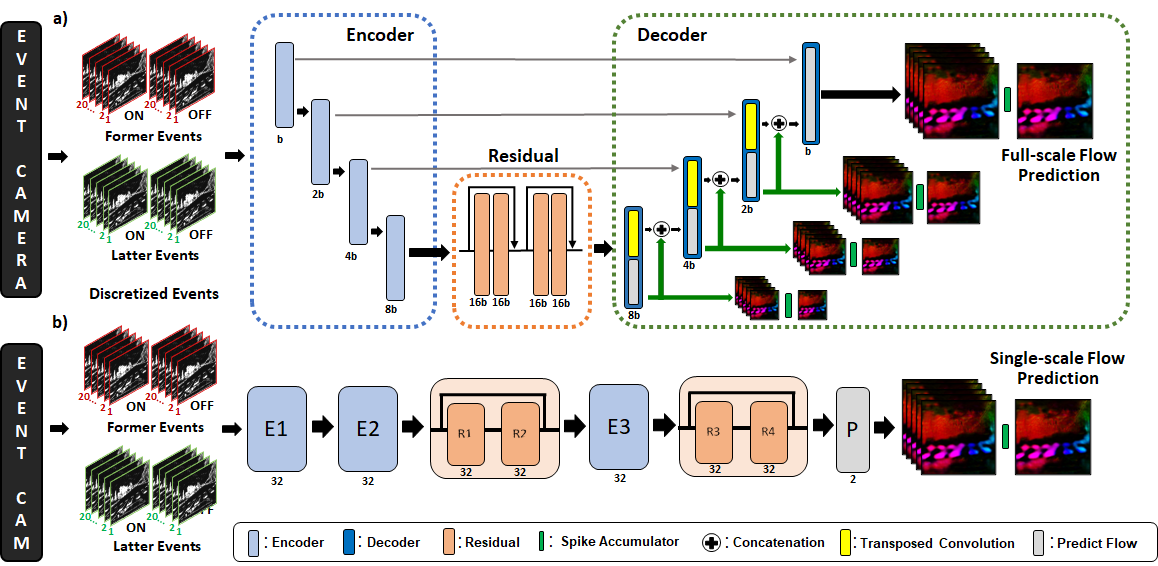}
\vspace{-1mm}
\caption{Fully-spiking architectures based on a) U-net~\cite{unet} and b) Fire-FlowNet~\cite{gehrig2021raft}. Best viewed in color.} 
\label{fig:arch}
\vspace{-7mm}
\end{center}
\end{figure*}

\subsection{Network Architecture} \label{subsec:arch}
We explore two types of fully-spiking architectures: an encoder-decoder based multi-scale architecture based on U-net~\cite{unet} and a lightweight single-scale architecture called Fire-FlowNet~\cite{paredes2021back}.
We evaluate several network sizes for the first architecture by subsequently scaling down the number of channels by a factor of $2$ at each layer. The Fire-FlowNet~\cite{paredes2021back} architecture performs no output downsampling and thus the input shape is maintained throughout the network. 

A forward pass involves passing the 4-channeled input representation (former and latter groups) sequentially over $T$ timesteps. At each timestep, the membrane potentials are updated (Eq.~\ref{eq4}) and output spikes are generated (Eq.~\ref{eq5}). The binary activations at every layer in the first type of architecture are downsampled/upsampled at each timestep before being forwarded to the next layer. The decoder layers upsample the provided input and also produce intermediate multi-scale flow predictions for each timestep. The flow prediction layers accumulate the flow over all timesteps to generate the final $Tanh$ activated full-scale flow prediction. Fire-FlowNet involves a similar forward pass but lacks any upsampling/downsampling layers. These architectures and discussed operations are illustrated in Fig.~\ref{fig:arch}. 

\subsection{Self-supervised Loss} \label{subsec:ssl_loss}
The Multi-Vehicle Stereo Event Camera Dataset (MVSEC)~\cite{zhu2018multivehicle} used in this work lacks reliable ground-truth labels. Thus, we adopt a self-supervised approach for training on this dataset~\cite{jason2016back}. The overall loss function is:
\begin{equation}
L^{\text{u}} = l_{\text{photo}} + \alpha l_{\text{smooth}} \tag{6}
\end{equation}
where $l_{photo}$ and $l_{smooth}$ represent photometric and, smoothness loss, respectively, and $\alpha$ denotes the weighting factor for smoothness loss.

\subsubsection{Photometric Loss} \label{subsubsec:photoloss}
The photometric loss imposes the brightness consistency assumption -- the intensity moving from pixel location $(x,y)$ at time $t$ to pixel location ($x+dx, y+ dy)$ at time $(t+ dt)$ remains the same. It is computed using two consecutive grayscale images $(I_t(x,y)$, $I_{t+dt}(x,y))$ and the predicted optical flow ($u,v$). A spatial transformer \cite{jaderberg2015} is used to inversely warp the second grayscale image $(I_{t+dt}(x,y))$ using the predicted optical flow ($u,v$). The photometric loss then minimizes the difference between the first grayscale image and the warped image as follows:
\begin{equation}
l_{photo} = \sum_{x,y} \rho(I_t(x,y) - I_{t+dt}(x+u, y+v)) \\ \tag{7}
\end{equation}
where $\rho$ is the robust Charbonnier loss $\rho(x) = (x^2 + \eta^2)^r$ used for outlier rejection \cite{sun2014}. We set $r$=$0.45$ and $\eta$=$1e^{-3}$.

\subsubsection{Smoothness Loss} \label{subsubsec:smoothloss}
Smoothness loss reduces the erratic variations in optical flow predictions at the edges by adding a regularization as follows:
\begin{multline}
    l_{smooth}  = \sum_{j} \sum_{i} (\|u_{i,j}-u_{i+1,j}\|  + \|u_{i,j}-u_{i,j+1}\| \\ + \|v_{i,j}-v_{i+1,j}\| + \|v_{i,j}-v_{i,j+1}\|) \tag{8}
\end{multline}

    
\vspace{2mm}
\noindent where $u_{i,j}$ and $v_{i,j}$ are the flow estimates at pixel location ($i,j$) in $x$ and $y$ directions, respectively.

\subsection{Supervised Loss} \label{subsec:supervised_loss}
We also evaluate our proposed architectures on the DSEC-Flow dataset~\cite{gehrig2021dsec, gehrig2021raft} which has much better ground-truth flow maps compared to the MVSEC dataset. Hence, we perform supervised learning on this dataset. Given the predicted optical flow ($u_{pred}, v_{pred}$) and the ground-truth optical flow ($u_{gt}, v_{gt}$), the total loss $L^{\text{s}}$ is computed as the mean squared error  (MSE) as follows:
\begin{equation}
    L^{\text{s}} = \frac{1}{n} \sum_{i=1}^K(u_{pred} - u_{gt})^2 + \frac{1}{n} \sum_{i=1}^K(v_{pred} - v_{gt})^2 \\ \tag{9}
\end{equation}
where $K$ is the number of pixels with non-zero flow. 

\begin{figure*}[ht]
\vspace{1.5mm}
\begin{center}
\includegraphics[width=0.85\textwidth]{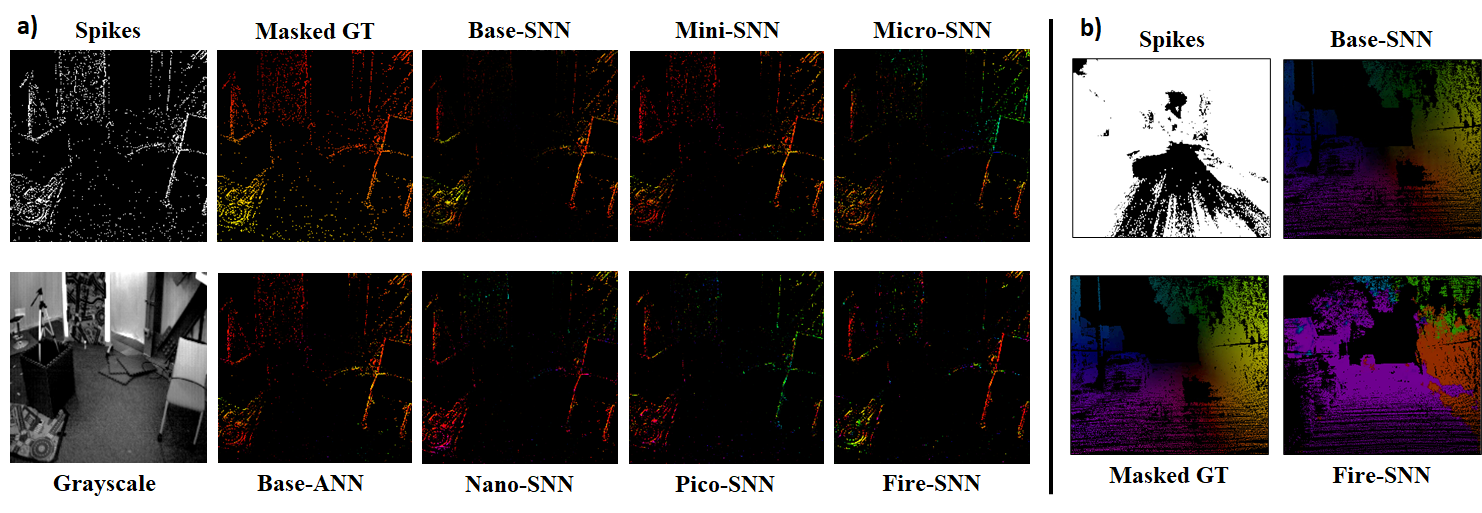}
\vspace{-1mm}
\caption{a) Results on MVSEC using various SNN architectures. b) Results on DSEC-Flow Base-SNN and Fire-SNN.}
\label{fig:results}
\end{center}
 \vspace{-4mm}
\end{figure*}

\subsection{Surrogate gradient and Backpropagation-through time} \label{subsec:backprop}
Once the final loss ($L$) (in both self-supervised and supervised cases) is obtained, the next step is to perform backpropagation and compute the loss gradients with respect to the network parameters. However, unlike ANNs, gradient computation in SNNs is not straightforward.

The spike generation mechanism of an LIF neuron results in a hard threshold function which is non-differentiable. Thus, there is a need to approximate its gradient using a surrogate function. We use the inverse tangent function~\cite{fang2021incorporating} as an approximation to the gradient of the LIF neuron, since it is widely used and is computationally inexpensive.
\[\pderiv{o^t_l}{z^t_l} =\begin{cases}
               \frac{1}{1+\gamma {z^t_l}^2}, & \text{if~$1-|z^t_l|>0$}\\ \tag{10}
               0, & \text{otherwise.}
            \end{cases} \]
            
\[\pderiv{o^t_l}{u^t_l} = \pderiv{o^t_l}{z^t_l} \pderiv{z^t_l}{u^t_l} = \pderiv{o^t_l}{z^t_l} \frac{1}{v^{th}_l}\\ \tag{11}
\]
The network also needs to be unrolled in time for all timesteps and the errors ($\pderiv{L}{o_{l}}$) need to be backpropagated using the surrogate gradient and Backpropagation Through Time (BPTT) \cite{werbos1990backpropagation}. Once all the gradients are obtained, the weight update can computed as the sum of gradients over each time-step. The weight updates for the SNN layer $l$ are described as follows:
\begin{equation}
\triangle \boldsymbol{W}_{l} = \sum_{t} \pderiv{\text{L}}{o^t_l} \pderiv{o^t_l}{z_l^t} \pderiv{z^t_l}{u_l^t}  \pderiv{u_l^t}{w_l} = \sum_{t} \pderiv{\text{L}}{o^t_l} \pderiv{o^t_l}{z_l^t} \frac{1}{v^{th}_l} o^t_{l-1}\\ \tag{12}
\end{equation}
Similarly, the threshold and leak updates are given by:
\begin{equation}
\triangle v^{th}_{l} = \sum_{t} \pderiv{\text{L}}{o^t_l} \pderiv{o^t_l}{z_l^t} \pderiv{z_l^t}{v^{th}_l} = \sum_{t} \pderiv{\text{L}}{o^t_l} \pderiv{o^t_l}{z_l^t} \left(\frac{-v^{th}_l o_l^{t-1} - u_l^t}{(v_l)^2} \right)\\ \tag{13}
\end{equation}
\begin{equation}
\triangle \lambda_{l} = \sum_{t} \pderiv{\text{L}}{o^t_l} \pderiv{o^t_l}{u_l^t} \pderiv{u_l^t}{\lambda_l} =  \sum_{t} \pderiv{\text{L}}{o^t_l} \pderiv{o^t_l}{u_l^t} u_l^{t-1}\\ \tag{14}
\end{equation}

We use layer-wise learnable thresholds and leaks which lead to a negligible increase in the model size. Using per channel thresholds and leaks does not lead to a significant performance improvement as highlighted in ~\cite{rathi2021diet}.
\section{Experiments} \label{sec:experiments}

\subsection{Datasets and Training Details} \label{subsec:dataset_details}
\subsubsection{MVSEC}\label{subsec:mvsec}
The MVSEC~\cite{zhu2018multivehicle} dataset contains four indoor flying, two outdoor day driving and, three outdoor night driving sequences. We perform training in SSL fashion using loss functions discussed in Section.~\ref{subsec:ssl_loss} on the $outdoor\_day2$ driving sequence and evaluate on $outdoor\_day1$ and $indoor\_flying1,2,3$ sequences for fair comparisons with past works~\cite{zhu2018ev, zhu2019unsupervised, lee2020_spikeflownet, lee2022fusion}. We also compare with other works ~\cite{paredes2021back, hagenaars2021self} that train on a separate high speed dataset (UZH-FPV Drone racing dataset~\cite{delmerico2019we}). However, these comparisons are not totally fair due to the drone racing dataset~\cite{delmerico2019we} having a much wider distribution of optical flow vectors than MVSEC leading to better learning. We train and evaluate on an event volume corresponding to a consecutive pair of grayscale images (dt=1).

During training, we perform random flips, rotations, and crop inputs to a $256 \times 256$ size. We use the Adam optimizer~\cite{kingma2015adam} and train for 100 epochs with a batch size of $8$ and an initial learning rate of $10^{-4}$. The learning rate is scaled by $0.7$ every $10$ epochs. The surrogate width $\gamma$ for the inverse tangent in the LIF neuron is set to $10$ and the smoothness loss weighting factor is also set to $10$. The number of bins ($B$) was set to $10$, leading to $5$ timesteps ($T$) and approximately $2000$ events per bin on average. These hyperparameters were chosen heuristically and based on past works.

\subsubsection{DSEC-Flow}\label{subsec:dsec}
The DSEC-Flow dataset was released in conjunction with~\cite{gehrig2021raft} adding to the original DSEC dataset~\cite{gehrig2021dsec}. It features VGA resolution event cameras and provides optical flow ground truths for 24 driving sequences for a total of 7800 training samples and 2100 test samples. However, the ground truths for the test sequences are not open-source, leaving us with only the training data. Thus, we create our own $80-20 (\%)$ split for training and testing using the available $7800$ samples corresponding to $18$ sequences. We train for $50$ epochs with random flips and a $288\times384$ crop. The learning rate is kept fixed at $10^{-4}$. 

Flow is evaluated using the average endpoint error (AEE) (also termed as endpoint error (EPE)) that measures the mean distance between the predicted and ground truth flow. This is done only for pixels containing events ($m$).
\begin{equation}
    AEE = \frac{1}{m}\sum_m \left\Vert(u,v)_{\text{pred}} - (u,v)_{\text{gt}}\right\Vert_2 \tag{15}
\end{equation}
\vspace{-4mm}

\begin{table*}[ht]
\vspace{1.5mm}
\caption{AEE results on MVSEC dataset. 1$^{st}$ block: Mixed SOTA methods. 2$^{nd}$ block: SSL methods using events and/or frames. 3$^{rd}$ block: Our ANN models. 4$^{th}$ block: Our fully-spiking SNNs. (E) Event-warping loss, (I) Photometric loss.}
 \vspace{-3mm}
\begin{center}
\resizebox{0.8\textwidth}{!}{
\begin{tabular}{lccccccc}
\hline \hline

 & Loss Type & outdoor\_day1 & indoor1 & indoor2 & indoor3 & AEE Sum & Improvement (\%)\\ \hline \vspace{-3mm} \\
EV-FlowNet~\cite{zhu2018ev} & I & 0.49 & 1.03 & 1.72 &  1.53 & 4.77 & -\\  \vspace{-3mm} \\
ECN$_\text{masked}$~\cite{ye2020_unsupervised} & E & \bf{0.30} & - & - &  - & - & -\\  \vspace{-3mm} \\
Zhe et. al~\cite{zhu2019unsupervised} & E & 0.32 & \ul{0.58} & \ul{1.02} &  \ul{0.87} & \bf{2.79}  & -\\  \vspace{-2.5mm} \\
Back to Event Basics$_{Evf}$~\cite{paredes2021back} & E & 0.92 & 0.79 & 1.40 &  1.18 & 4.29 & - \\   \vspace{-3mm} \\
Back to Event Basics$_{Fire}$~\cite{paredes2021back} & E & 1.06 & 0.97 & 1.67 &  1.43 & 5.13 & -\\  \vspace{-3mm} \\
XLIF-EV-FlowNet~\cite{hagenaars2021self} & E & 0.45 & 0.73 & 1.45 &  1.17 & 3.8 & -\\  \vspace{-3mm} \\
XLIF-FireNet~\cite{hagenaars2021self} & E & 0.54 & 0.98 & 1.82 &  1.54 & 4.88  & -\\ \hline  \vspace{-3mm} \\
Spike-FlowNet~\cite{lee2020_spikeflownet} & I & 0.49 & 0.84 & 1.28 &  1.11 & 3.72 & - \\  \vspace{-3mm} \\
Fusion-FlowNet~\cite{lee2022fusion} & I & 0.59 & \bf{0.56} & \bf{0.95} &  \bf{0.76} & \ul{2.86} & -\\ \hline  \vspace{-3mm} \\

Ours (Base-ANN) & I & 0.48 & 0.84 & 1.59 &  1.36 & 4.27 & -\\  \vspace{-3mm} \\
Ours (Mini-ANN) & I & 0.52 & 0.9 & 1.68 &  1.44 & 4.54 & -\\  \vspace{-3mm} \\
Ours (Micro-ANN) & I & 0.57 & 0.95 & 1.74 &  1.48 & 4.74 & -\\  \vspace{-3mm} \\
Ours (Nano-ANN) & I & 0.62 & 0.97 & 1.74 &  1.49 & 4.82 & -\\  \vspace{-3mm} \\
Ours (Pico-ANN) & I & 0.65 & 1.04 & 1.78 &  1.53 & 5 & -\\  \vspace{-3mm} \\
Ours (Fire-ANN) & I & 1.01 & 1.22 & 2.03 &  1.78 & 6.04 & -\\ \hline \vspace{-3mm} \\

Ours (Base-SNN) & I & \ul{0.44} & 0.79 & 1.37 &  1.11 & 3.71 & 13.1\\  \vspace{-3mm}\\
Ours (Mini-SNN) & I & 0.46 & 0.83 & 1.4 &  1.17 & 3.86 & \ul{14.9}\\  \vspace{-3mm} \\
Ours (Micro-SNN) & I & 0.52 & 0.92 & 1.53 &  1.28 & 4.25 & 10.3\\  \vspace{-3mm} \\
Ours (Nano-SNN) & I & 0.52 & 0.93 & 1.54 & 1.31 & 4.3 & 10.7\\  \vspace{-3mm} \\
Ours (Pico-SNN) & I & 0.58 & 0.95 & 1.58 &  1.28 & 4.39 & 12.2\\  \vspace{-3mm} \\
Ours (Fire-SNN) & I & 0.72 & 1.08 & 1.8 &  1.47 & 5.07 & \bf{16.1}\\ \hline \hline \vspace{-3mm}
\end{tabular}}
\end{center}
\vspace{-4mm}
\label{table1}
\end{table*}

\begin{table*}[ht]
\caption{Computational efficiency on MVSEC \small{\textit{(* $\#$OPS for corresponding ANN)}}}
\vspace{-3mm}
\begin{center}
\resizebox{0.9\textwidth}{!}{
\begin{tabular}{lcccccc}
\hline \hline
 & $\# \text{Parameters} (\times 10^6)$ & $\# \text{OPS}_\text{ANN}(\times 10^9)$ & $\text{Avg. Spiking Activity} (\%)$ & $\# \text{OPS}_\text{SNN}(\times 10^9)$ & $\text{E}_\text{Total} (mJ)$ & Improvement ($\times$)\\ \hline \vspace{-3mm} \\
Ev-FlowNet~\cite{zhu2018ev} & 13.04 & 5.34 & - & - &  24.54 & 1$\times$ \\  \vspace{-3mm} \\
Spike-FlowNet~\cite{lee2020_spikeflownet} & 13.04 & 4.41 & 0.48 & 0.0158 & 20.29 & 1.21$\times$\\ \vspace{-3mm} \\
Fusion-FlowNet$_{\text{Late}}$~\cite{lee2022fusion}  & 7.55 & 2.85 & 0.147 & 0.0052 & 13.11 & 1.87$\times$ \\ \hline \vspace{-3mm} \\
Ours (Base-SNN) & 13.04 & 11.31* & \ul{45.83} & 25.92 & 23.3 & 1.05$\times$ \\ \vspace{-3mm} \\
Ours (Mini-SNN) & 3.41 & 4.3* & 45.15& 9.71 & 8.75 & 2.8$\times$ \\ \vspace{-3mm} \\
Ours (Micro-SNN) & 0.93 & 1.88* & 61.9& 5.81 & 5.25 & 4.66$\times$ \\ \vspace{-3mm} \\
Ours (Nano-SNN) & 0.27 & \ul{0.96*} & 55.55 & \ul{2.67} & \ul{2.4} & \ul{10.2}$\times$\\ \vspace{-3mm} \\
Ours (Pico-SNN) & \ul{0.092} & \bf{0.57*} & 74.12& \bf{2.11} & \bf{1.9} & \bf{12.9}$\times$ \\ \vspace{-3mm} \\
Ours (Fire-SNN) & \bf{0.057} & 3.7* & \bf{28.57} & 5.28 & 4.75 & 5.16$\times$ \\ \hline \hline \vspace{-3mm} 
\end{tabular}}
\end{center}
\vspace{-6mm}
\label{table2}
\end{table*}

\subsection{Architectural Ablations}
We analyze several smaller model sizes by reducing the number of base channels (b) in the original U-Net architecture. Namely the architectures are Base(64), Mini(32), Micro(16), Nano(8) and Pico(4), where the quantity in parentheses represents the base channels (b) for that model. All these architectures along with Fire-FlowNet-SNN are trained in an SSL fashion on the MVSEC dataset and in a supervised manner on the DSEC-Flow dataset. In addition, we also train their corresponding ANN models to have an iso-architecture comparison between SNN and ANN implementations. Note, that the SNN and ANN models are based on architectures explored in ~\cite{lee2020_spikeflownet} and ~\cite{zhu2018ev}, respectively.  For the ANN training, the event bins are passed as channels rather than as a sequence. Table.~\ref{table3} shows the number of parameters in each model. Note that, even though Fire-FlowNet has the least number of parameters ($~57K$), it incurs a hefty $3.7 \times 10^9$ operations due to no downsampling of activations.

\begin{table}[ht]
\caption{EPE results on DSEC-Flow dataset.}
 \vspace{-3mm}
\begin{center}
\resizebox{0.36\textwidth}{!}{
\begin{tabular}{lcccccc}
\hline \hline
 & EPE & 1PE & 2PE & 3PE \\ \hline \vspace{-3mm} \\
EV-FlowNet~\cite{zhu2018ev} & 2.32 & 55.4 & 29.8 & 18.6 \\  \vspace{-2.5mm} \\
E-RAFT~\cite{gehrig2021raft} & \bf{0.79} & \bf{12.5} & \bf{4.7} & \bf{2.7}\\ \hline  \vspace{-2.5mm} \\
Ours (Base-ANN) & 1.82 & 22.5 & 9.8 & 5.1 \\  \vspace{-2.5mm} \\
Ours (Base-SNN) & \ul{1.62} & \ul{19.2} & \ul{8.4} & \ul{4.5} \\ \hline \vspace{-2.5mm} \\
Ours (Fire-ANN) & 5.92 & 82.6 & 59.1 & 40.2 \\  \vspace{-2.5mm} \\
Ours (Fire-SNN) & 4.88 & 66.8 & 37.5 & 24.9\\ \hline \hline \vspace{-2.5mm}
\end{tabular}}
\end{center}
\vspace{-8mm}
\label{table3}
\end{table}

\subsection{Quantitative Results} \label{subsec:results}
The quantitative results are showcased in Table.~\ref{table1} and Table.~\ref{table3} and visualized in Fig.~\ref{fig:results}. From Table~\ref{table1} we observe that Fusion-FlowNet outperforms all methods for nearly all sequences. This is mainly due to the usage of frame information in addition to events. The second best is~\cite{zhu2019unsupervised}, the unsupervised pipeline trained on MVSEC using the event-warping (E) loss. Our ANN models demonstrate respectable performance compared to prior works, and show a gradual increase in AEE with decreasing model size. The result of interest is that our fully-spiking models consistently outperform the corresponding ANN models ($13\%$ lower AEE on average) for all model sizes. In fact, our Base-SNN shows similar or better performance compared to the only other fully-spiking work (XLIF-EV-FlowNet~\cite{hagenaars2021self}) even with the inferior intensity based (I) loss. Although, the corresponding FireNet model (XLIF-FireNet~\cite{hagenaars2021self}) marginally outperforms our Fire-SNN model. This can be well attributed to the event-warping (E) loss and also the use of FireNet which contains explicit recurrence in the form of GRU units.

On DSEC-Flow, our work, although unable to outperform E-RAFT~\cite{gehrig2021raft} which again uses explicit recurrence and iterative refinement, clearly outclasses EV-FlowNet~\cite{zhu2018ev} by $\sim30\%$. E-RAFT is also evaluated on the original DSEC-Flow testset. Again, fully-spiking models show lower EPE $\sim11\%$ (Base) and  $\sim17\%$ (Fire) compared to ANN models.

\subsection{Computational Efficiency} \label{subsec:energy}
We validate the efficiency of our fully-spiking models in terms of the number of network parameters and inference energy. For estimating the inference energy, we utilize the estimation method used in~\cite{lee2022fusion, rueckauer2017conversion}. SNNs perform sparse ACcumulate (AC) operations due to their binary outputs, whereas ANNs perform dense Multiply-and-ACcumulate (MAC) operations.
The energy required for MAC and AC operations are: $E_{MAC}$=$4.6pJ$ and $E_{AC}$=$0.9pJ$, for a 32-bit floating-point computation in 45nm CMOS technology~\cite{horowitz20141}. This makes the AC operation $5.1 \times$ more energy-efficient than MAC. The layer-wise synaptic operations in SNNs can be computed as the product of the \#neurons ($M_l$), mean firing activity ($F_l$), \#synaptic connections ($C_l$) and, \#timesteps ($T$).
\begin{equation}
\text{\#OPS}_{\text{SNN}} =  T\sum_{l}M_l  C_l F_l, \;
\text{\#OPS}_{\text{ANN}} = \sum_{l} M_l C_l\\\tag{16}
\label{eq12}
\end{equation}
\vspace{-8mm}

\begin{equation}
\text{E}_{\text{SNN}} = \text{\#OPS}_{\text{SNN}} \times \text{E}_{\text{AC}}, \;\;\;\;\;
\text{E}_{\text{ANN}} = \text{\#OPS}_{\text{ANN}} \times \text{E}_{\text{MAC}}\\ \tag{17}
\label{eq13}
\end{equation}

Table ~\ref{table2} shows the computational efficiency of our proposed SNNs. Our SNN models show much higher spiking activity compared to hybrid SNN-ANNs~\cite{lee2020_spikeflownet, lee2022fusion}. This is because the models learn to reduce the LIF thresholds and increase leak to overcome vanishing spikes, allowing for better performance while still being efficient. Our SNNs also incur much lower computational cost due to sparse event-driven and binary computations offering $1-2$ orders of magnitude improvement in compute energy. We observe that Nano-SNN offers an optimal tradeoff between performance ($10.7\%$ and $9.9\%$ lower AEE) and efficiency ($1.84\times$ and $10.2\times$ lower energy) than Nano-ANN and \cite{zhu2018ev}, respectively.

\section{CONCLUSION}
 We propose a framework for training deep fully-spiking networks for optical flow estimation using adaptive neuronal dynamics. We show that our models capture  timing information better making them suitable as low-latency and low-energy solutions for the resource-constrained edge while maintaining application accuracy. This work can be extended to other perception tasks, eventually fueling the controls and planning pipelines and enabling high-speed and  collision-free robot navigation.
\newpage

\bibliography{main.bbl} 
\bibliographystyle{ieeetr}

\end{document}